IAC–24–D1,IP,39,x87432

# DreamSat: Towards a General 3D Model for Novel View Synthesis of Space Objects

**Nidhi Mathihalli[a], Audrey Wei[b], Giovanni Lavezzi[c], Peng Mun Siew[d], Victor Rodriguez-Fernandez[e], Hodei Urrutxua[f], Richard Linares[g]**

[a] *Undergraduate student, Department of Aeronautics and Astronautics, Massachusetts Institute of Technology, MA 02139, USA*, nidhim27@mit.edu
[b] *Undergraduate student, Department of Aeronautics and Astronautics, Massachusetts Institute of Technology, MA 02139, USA*, weia@mit.edu
[c] *Postdoctoral Associate, Department of Aeronautics and Astronautics, Massachusetts Institute of Technology, MA 02139, USA*, glavezzi@mit.edu
[d] *Research Scientist, Department of Aeronautics and Astronautics, Massachusetts Institute of Technology, MA 02139, USA*, siewpm@mit.edu
[e] *Associate Professor, School of Computer Systems Engineering, Universidad Politécnica de Madrid*, victor.rfernandez@upm.es
[f] *Associate Professor, School of Aerospace Engineering, Universidad Rey Juan Carlos*, hodei.urrutxua@urjc.es
[g] *Associate Professor, Department of Aeronautics and Astronautics, Massachusetts Institute of Technology, MA 02139, USA*, linaresr@mit.edu
[a,b] The authors N. Mathihalli and A. Wei have equally contributed to this work.

## Abstract

Novel view synthesis (NVS) enables to generate new images of a scene or convert a set of 2D images into a comprehensive 3D model. In the context of Space Domain Awareness, since space is becoming increasingly congested, NVS can accurately map space objects and debris, improving the safety and efficiency of space operations. Similarly, in Rendezvous and Proximity Operations missions, 3D models can provide details about a target object's shape, size, and orientation, allowing for better planning and prediction of the target's behavior. In this work, we explore the generalization abilities of these reconstruction techniques, aiming to avoid the necessity of retraining for each new scene, by presenting a novel approach to 3D spacecraft reconstruction from single-view images, DreamSat, by fine-tuning the Zero123 XL, a state-of-the-art single-view reconstruction model, on a high-quality dataset of 190 high-quality spacecraft models and integrating it into the DreamGaussian framework. We demonstrate consistent improvements in reconstruction quality across multiple metrics, including Contrastive Language-Image Pretraining (CLIP) score (+0.33%), Peak Signal-to-Noise Ratio (PSNR) (+2.53%), Structural Similarity Index (SSIM) (+2.38%), and Learned Perceptual Image Patch Similarity (LPIPS) (+0.16%) on a test set of 30 previously unseen spacecraft images. Our method addresses the lack of domain-specific 3D reconstruction tools in the space industry by leveraging state-of-the-art diffusion models and 3D Gaussian splatting techniques. This approach maintains the efficiency of the DreamGaussian framework while enhancing the accuracy and detail of spacecraft reconstructions. The code for this work can be accessed on GitHub (https://github.com/ARCLab-MIT/space-nvs).

## 1. Introduction

Three-dimensional (3D) reconstruction from images has become a pivotal area of research in computer vision, with significant implications for various fields, including space exploration. This process aims to generate 3D models from two-dimensional (2D) images, enabling a more comprehensive understanding of objects and environments. Image-to-3D reconstruction is particularly valuable in the space domain for its many applications, improving the effectiveness, quality, and cost of mission planning, navigation and maneuvers, virtual simulations, remote diagnostics of spacecraft in orbit, spacecraft design, debris monitoring, and risk assessments.

Recent advancements in deep learning and computer vision have led to remarkable progress in this domain. In addition, the rise of new 3D reconstruction techniques including Neural Radiance Fields (NeRF) [1] and 3D Gaussian splatting (3DGS) [2] have significantly improved the accuracy and quality of 3D reconstructions. NeRF-based methods use fully-connected deep networks to model volumetric scene functions from a sparse set of viewpoints. Some variants of NeRF are *NeRF–*, with no known camera parameters [3], Instant NeRF [4] with faster training times, and Mip-NeRF360 [5] with improved rendering efficiency and reduced aliasing artifacts across a 360° view. Regarding generalization capabilities of NeRF, scene-specific pre-training, meta-learning, and other techniques are often used [6]. Pre-training allows the model to train on images that are close to the desired datasets or point clouds, aligning weights to improve model performance. Meta-learning involves changing weights or tuning models to focus on fewer examples, allowing it to adapt to a diverse dataset.

While NeRF produces impressive results, it suffers from slow rendering times and requires multiple views. 3DGS offers faster rendering and optimization. 3DGS involves Structure-from-Motion, a photogrammetric technique to create 3D structures from 2D images that produces a sparse point cloud and uses splatting techniques to render novel views of the scene. 3DGS achieves comparable training times to Instant NeRF and quality to Mip-NeRF360. However, 3DGS still faces challenges in texture accuracy and computational efficiency, as well as still requiring multiple images from different viewpoints.

The rise of large-scale 3D model datasets like Objaverse [7] has enabled significant progress in single-view 3D reconstruction, which only requires a single view to generate a 3D model of an object. This advancement has led to the development of several notable approaches. Zero123 [8] leverages a diffusion-based generative model to predict novel views from a single input image. It demonstrates impressive generalization across various object categories but may struggle with complex, out-of-distribution shapes. Zero123++ [9] builds upon Zero123 by incorporating additional priors and refinement techniques, resulting in improved reconstruction quality and consistency. SynchDreamer [10] introduces a synchronous generation framework that simultaneously produces multiple views, leading to more coherent 3D reconstructions. Zero123XL extends Zero123 by training on the expanded ObjaverseXL dataset [11] and employing alignment finetuning techniques. This results in enhanced performance across a broader range of object categories.

Building on these single-view reconstruction methods, several approaches have emerged to generate complete 3D assets by incorporating diffusion processes to improve the fidelity of generated images. DreamFusion [12] combines NeRF with text-to-image diffusion models to generate 3D objects from text descriptions. While innovative, it suffers from long optimization times. DreamGaussian [13] adapts 3D Gaussian Splatting for generative tasks, significantly reducing optimization time compared to NeRF-based methods. It also introduces efficient mesh extraction and texture refinement techniques.

Despite these advancements, a significant gap remains in the field of spacecraft-specific 3D reconstruction. Current general-purpose datasets like Objaverse have limited representation of spacecraft and so do not fully encompass the diverse structural characteristics of spacecraft, limiting the learning potential of these models. Thus, a major challenge remains: the lack of specialized single-view reconstruction models for the use of spacecraft. Unlike previous studies that focus on generic image datasets, this research specifically targets spacecraft models, aiming to improve the model's performance in this domain.

The paradigm shift towards generalized models represents a pivotal advancement in Artificial Intelligence (AI), particularly in the realm of 3D modeling. This novel approach, which we are pioneering for the space domain, marks a significant departure from traditional methodologies. Unlike the prevalent practice of fine-tuning Large Language Models (LLMs) for each specific domain, our research focuses on the innovative concept of fine-tuning 3D models. This strategy not only streamlines the adaptation process but also potentially unlocks new capabilities in space-related applications. To the best of our knowledge, DreamSat is not only the first of its kind to effectively finetune such a model to be used in the field of space, but the first to finetune for any specific domain, opening the doors to applying such methods for use in any field. By pursuing this direction, we aim to demonstrate the versatility and efficiency of generalized 3D models, potentially revolutionizing how we approach domain-specific AI applications in space exploration and related fields.

To address this gap, we present a novel approach that finetunes the state-of-the-art Zero123XL model on a curated dataset of high-quality spacecraft images within the DreamGaussian pipeline. This fine-tuning process within the DreamGaussian framework allows for efficient, high-quality, and texture-rich 3D reconstruction specifically tailored to spacecraft, resulting in superior reconstruction performance than the baseline DreamGaussian method. By leveraging the strengths of Zero123XL and DreamGaussian, this specialized 3D reconstruction framework for spacecraft achieves improved reconstruction quality and efficiency across the board, as demonstrated by improvements of Contrastive Language-Image Pretraining (CLIP) similarity, Peak Signal-to-Noise Ratio (PSNR), Structural Similarity Index (SSIM), and Learned Perceptual Image Patch Similarity on spacecraft imagery. By focusing on this domain, we enable more accurate and reliable 3D modeling, while maintaining the efficiency of

DreamGaussian.

This paper is organized as follows: Section 2 covers the fundamentals of the dataset, models, and metrics. Section 3 delves into the contributions of the model. Section 4 presents the results, followed by a discussion. Finally, Section 5 concludes the paper.

## 2. Fundamentals

### 2.1 The Objaverse Dataset

The Objaverse [7] dataset comprises over 800,000 3D assets. While this dataset does have a diverse collection of different objects, it lacks volume in specific categories. For example, out of 817,899 total objects in the Objaverse dataset, less than 1000 images are of spacecraft, space shuttles, or satellites, making them severely underrepresented. Furthermore, the spacecraft models within Objaverse lack diversity, as the majority are basic rocket models.

### 2.2 Zero123

Zero123 [8] is a diffusion-based generative model designed for single-view 3D reconstruction based on an encoder-decoder architecture. The encoder extracts features from the input image, and the decoder uses these features to predict a latent 3D representation. This latent code is then decoded into a final 3D mesh representing the object in the image. The key components of Zero123 include a conditional diffusion model that generates novel views based on the input image and target camera pose, a view-conditioned score distillation sampling (SDS) process that optimizes the 3D representation, and a multi-view consistency loss that ensures coherence across generated views. As Zero123 demonstrates remarkable generalization capabilities across various object categories, it is used as the foundation for specialized 3D reconstruction tasks.

### 2.3 Zero123XL

As part of our DreamSat pipeline, we integrate and finetune Zero123XL model [11]. Zero123XL builds upon the original Zero123 model with several key enhancements. It is trained on ObjaverseXL, an expanded dataset containing a more diverse range of over 10 million 3D objects, enabling better generalization. It also has undergone alignment finetuning, which is a technique that aligns the model's outputs more closely with desired characteristics.

### 2.4 Alignment Finetuning

Alignment fine-tuning is a technique used to adapt pretrained models to specific domains or tasks while maintaining their general capabilities. This approach has been successfully applied in various contexts including Human Preference Alignment [14], which is fine-tuning language models to align with human preferences and values, Domain Specialization [15], adapting general models to perform well on specific domains, such as medical, financial, or legal text, and Task-Specific Alignment [16], which is fine-tuning models for particular tasks while preserving general knowledge.

In our case, alignment fine-tuning is used to specialize Zero123XL for spacecraft reconstruction to accurately reflect spacecraft shapes and textures, while retaining its general 3D modeling capabilities.

### 2.5 Dream Gaussian

The DreamGaussian framework [13] leverages generative 3D Gaussian splatting, which uses Gaussian functions to represent the 3D space of an object. This method provides a continuous representation, enhancing the fidelity and texture of the reconstructed images. The framework integrates the finetuned Zero123XL model as a 2D diffusion prior, which guides the generation process by refining the Gaussian noise into detailed images. The DreamGaussian pipeline first initialize a set of 3D Gaussians based on the input image. Then, it repeats the following process until convergence of rendering the Gaussians from multiple viewpoints, using the 2D diffusion prior (Zero123XL) to evaluate and improve the rendered views, and optimizing the 3D Gaussians based on the feedback from the diffusion prior. This approach allows for rapid 3D reconstruction while leveraging the power of large-scale 2D diffusion models.

### 2.6 Metrics

The metrics we will use to compare our results to existing ones are CLIP Similarity, PSNR, SSIM, and LPIPS. CLIP similarity measures the semantic similarity between generated and ground truth images. PSNR measures the quality of the created three-dimensional point cloud by measuring the mean-square error between the new and input — or in our use case, current — point clouds. The higher the PSNR, the higher the quality. SSIM measures local pixel patterns normalized for structural information based on luminance, contrast, and structure. It defers from PSNR in that in that it also measures perceived quality rather than only absolute error. The higher the SSIM, the more perfect the structural similarity. Lastly, we used LPIPS, which extracts features from both point clouds and measures a weighted average distance to measure the patch similarity. The smaller the distance, the closer the two point clouds; hence, the smaller LPIPS, the closer two point clouds are close to each other.

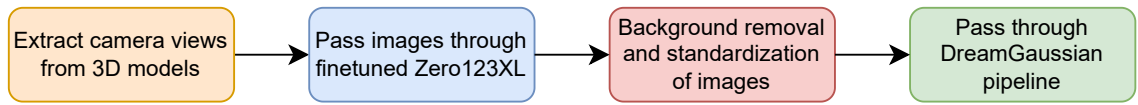

Figure 1: DreamSat pipeline

## 3. DreamSat

### 3.1 Model Architecture/Pipeline

By integrating the finetuned Zero123XL model into the DreamGaussian pipeline, we create a specialized framework for spacecraft 3D reconstruction that offers several advantages. We gain domain-specific knowledge, as the finetuned Zero123XL model captures the unique characteristics of spacecraft structures and textures. We also achieve efficient optimization, as DreamGaussian's 3D Gaussian splatting enables rapid convergence and high-quality results. In addition, we get high-fidelity textures, as the refinement stage enhances surface details crucial for accurate spacecraft representation. Also, the resulting textured mesh is readily usable in various 3D applications and analysis tools. This combined approach allows us to leverage the strengths of both models, resulting in a powerful tool for spacecraft 3D reconstruction from single images. The DreamSat pipeline is shown in Fig. 1.

### 3.2 Dataset Preparation and Training Methodology

We curated a dataset of 190 high-quality spacecraft models from National Aeronautics and Space Administration (NASA) [17], European Space Agency (ESA) [18], and Satellite Pose Estimation and 3D Reconstruction (SPE3R) [19, 20], which comprises 64 unique spacecraft 3D models, allowing for generalization across common spacecraft designs. This diverse set of models encompasses a wide range of spacecraft types, including satellites, space probes, and orbital stations.

To prepare the dataset for Zero123XL fine-tuning, we processed each 3D model (in .obj format) to extract 48 camera views. This was achieved by rotating the camera around XY, YZ, and XZ planes, extracting 16 views evenly spaced at 22.5° intervals per plane.

This approach ensures comprehensive coverage of each spacecraft from multiple angles, enabling the model to learn robust representations of spacecraft structures. Furthermore, employ a continual learning strategy. The data is divided into chunks, and the model is trained incrementally using checkpoints, making efficient use of GPU resources. We also allocate 20 spacecraft for our validation dataset.

### 3.3 Fine-tuning Zero123XL

We utilize the pre-trained Zero123XL checkpoint as a starting point, allowing the model to leverage its general understanding of 3D structures while adapting to the specific characteristics of spacecraft. Following the fine tuning implementation of Zero123-XL, we performed fine tuning through continual learning with a learning rate of 5e-5, batch size of 1, and 48 data chunks using 5 NVIDIA GeForce RTX 4090 GPUs, taking on average three minutes per cycle, so around 2.5 hours total.

### 3.4 Dream Gaussian - Image Processing

DreamGaussian performs background removal and standardizes images into a certain size to be used in training, while also converting images to RGBA (red green blue alpha) form.

### 3.5 Integration of Dream Gaussian with Finetuned Model

The finetuned Zero123XL model serves as the 2D diffusion prior within the DreamGaussian framework. We integrated our specialized model by inserting the finetuned checkpoint and corresponding configuration file into the DreamGaussian pipeline. This integration allows us to benefit from DreamGaussian's efficient 3D Gaussian splatting and texture refinement techniques while leveraging our spacecraft-specific knowledge. The rest of the DreamGaussian framework remained unchanged, maintaining its efficient reconstruction capabilities

## 4. Results and Discussion

Our fine-tuned Zero123XL model, integrated into the DreamGaussian framework, demonstrated improvements in 3D reconstruction quality for spacecraft images.

|  | PSNR ↑ | SSIM ↑ | LPIPS ↓ | CLIP ↑ |
| --- | --- | --- | --- | --- |
| Original | 0.921 | 0.033 | 0.598 | 0.457 |
| Finetuned | 0.944 | 0.034 | 0.597 | 0.458 |
| Improvement | **2.53%** | **2.38%** | **0.16%** | **0.33%** |

Table 1: Comparison of PSNR, SSIM, LPIPS, and CLIP between different methods.

As shown in Table 1, on a test set of 30 previously unseen NASA spacecraft images, we observed the follow-ing improvements: CLIP Score: +0.33%, PSNR: +2.53%, SSIM: +2.38%, LPIPS: +0.16%. These improvements indicate enhanced performance across semantic understanding (CLIP), image quality (PSNR), structural similarity (SSIM), and perceptual quality (LPIPS).

| Space Vehicle | Input | Generated Novel Views | | | |
|---|---|---|---|---|---|
| Explorer 1 | 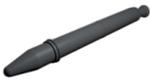 | 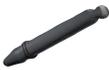 | 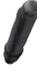 | 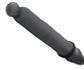 | 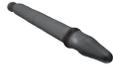 |
| Apollo Lunar Module | 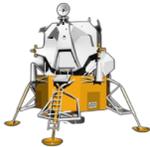 | 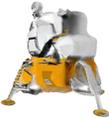 | 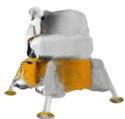 | 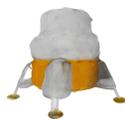 | 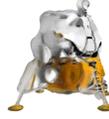 |
| Space Launch System Block 1 | 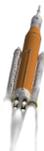 | 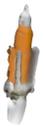 | 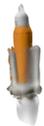 | 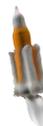 | 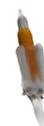 |
| High Energy Solar Spectroscopic Imager (HESSI) | 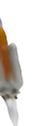 | 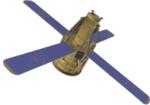 | 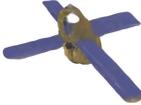 | 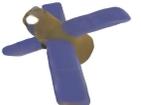 | 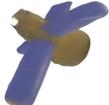 |
| Global Precipitation Measurement (GPM) | 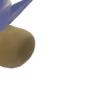 | 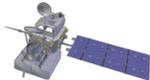 | 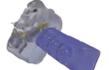 | 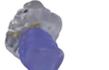 | 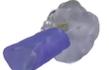 |
| Compton Gamma Ray Observatory (CGRO) | 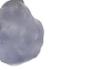 | 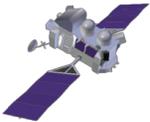 | 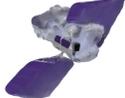 | 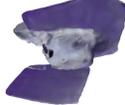 | 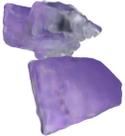 |
| Psyche Spacecraft | 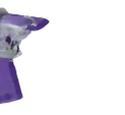 | 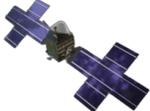 | 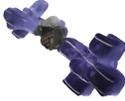 | 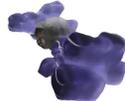 | 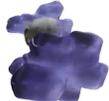 |

Table 2: Generated novel views of space vehicles

Table 2 provides a visual assessment of DreamSat's novel view synthesis capabilities, showcasing a collection of reconstructed views generated from single images of various space vehicles. Both successful examples that highlight the fidelity and detail of our approach, as well as instances where the model encountered challenges, illustrating limitations that warrant further investigation and refinement, are included.

Notably, the reconstruction time remained consistent with the original DreamGaussian framework, taking only a couple of minutes per reconstruction via MIT SuperCloud High Performance Computing cluster [21]. This efficiency is a significant advantage over other methods that may require hours or even days for high-quality 3D reconstruction. The improvements indicate that our specialized model, DreamSat, captures spacecraft-specific features more accurately than the general-purpose Zero123XL model.

### 4.1 Importance of 3D Spacecraft Reconstruction

The improved performance of our specialized spacecraft 3D reconstruction model has significant implications for the space domain such as:

- Efficiency in space mission planning: the rapid reconstruction time (minutes instead of hours) enables quick iterations in spacecraft design and mission planning processes, potentially reducing development cycles and costs.

- Remote analysis capabilities: high-quality 3D reconstructions from single images enhance the ability to analyze and assess spacecraft conditions remotely, which is crucial for ongoing missions and space debris management.

### 4.2 Scalability and Adaptability

Even with a relatively small dataset of 190 spacecraft models, we achieved noticeable improvements across all metrics. This suggests that scaling up the dataset size and diversifying the dataset could lead to even more significant enhancements, following the trend observed in the transition from Zero123 to Zero123XL.

In addition, while using single view input is convenient, in space there could be multiple pictures of the same object taken during a proximity operation. Adding more views of the same object may significantly improve the reconstruction, which is something that has potential to explore further.

Our results also demonstrate that general-purpose 3D reconstruction models can be effectively finetuned for specialized use cases, opening doors for adaptation to various domains within and beyond the space industry.

### 4.3 Limitations

However, we also identified some limitations and areas for future work. Firstly, there are data and computational constraints. The training of large models like Zero123XL is computationally intensive, requiring significant GPU resources. Expanding the dataset and model size may pose challenges for researchers with limited computational capabilities.

We also have the limitation of diversity in spacecraft shapes. Our model showed some difficulties in reconstructing more unusual or complex spacecraft geometries. Incorporating a more diverse range of spacecraft designs in the training set could address this limitation.

Another limitation regards fine-grained details. While overall reconstruction quality improved, capturing extremely fine details or small components of spacecraft remains challenging. Further research into high-resolution texture synthesis or multi-scale approaches could yield improvements in this area.

### 5. Conclusions

DreamSat demonstrates the potential of combining state-of-the-art 3D reconstruction techniques with domain-specific finetuning for spacecraft modeling. By adapting Zero123XL within the DreamGaussian framework, we achieved improved reconstruction quality and maintained high efficiency, contributing valuable tools to the space domain.

While our current results are promising, they also point to the potential for further improvements with larger datasets and more computational resources. The scalability of this approach suggests that continued research in this direction could lead to even more significant advancements in 3D reconstruction capabilities for the aerospace industry.

The success of this approach highlights the importance of specialized datasets and models in pushing the boundaries of 3D reconstruction for critical applications like Space Domain Awareness, rendezvous and proximity operations missions, space exploration, and satellite design. As larger and more diverse spacecraft datasets become available, we anticipate further improvements in reconstruction accuracy and generalization capabilities.

The high quality and efficiency of our method open new possibilities for rapid prototyping, mission planning,

and remote analysis in the space industry. By leveraging advancements in computer vision and applying them to space-specific challenges, we contribute to the ongoing evolution of space technology and exploration capabilities.

Future work should focus on studying how the size of the finetuning dataset affects the performance of the finetuned model, expanding the spacecraft dataset, exploring multi-scale reconstruction techniques, and investigating the integration of physics-based constraints to further enhance the fidelity of 3D reconstructions for space applications.

### Acknowledgments

Research was sponsored by the Department of the Air Force Artificial Intelligence Accelerator and was accomplished under Cooperative Agreement Number FA8750-19-2-1000. The views and conclusions contained in this document are those of the authors and should not be interpreted as representing the official policies, either expressed or implied, of the Department of the Air Force or the U.S. Government. The U.S. Government is authorized to reproduce and distribute reprints for Government purposes notwithstanding any copyright notation herein.

The authors acknowledge the MIT SuperCloud for providing HPC resources that have contributed to the research results reported within this paper.

H.U. wishes to acknowledge support through the research grant TED2021-132099B-C32 funded by MCIN/AEI/10.13039/501100011033 and the "European Union NextGenerationEU/PRTR".